\documentclass[12pt]{article}

\usepackage{fancyhdr}
\usepackage[]{graphicx}
\usepackage{caption}
\usepackage{subcaption}
\usepackage{times}
\usepackage[margin=1.0in]{geometry}
\usepackage{anyfontsize}
\usepackage{hyphenat}

\usepackage{etoolbox}
\patchcmd{\thebibliography}{\section*{\refname}}{}{}{}

\makeatletter
\renewcommand\@biblabel[1]{\textbullet}
\makeatother

\graphicspath{{figures/}} 

\begin{document}
{\fontsize{16}{19.2}\selectfont \textbf{PELVIS SURFACE ESTIMATION FROM PARTIAL CT FOR\\COMPUTER-AIDED PELVIC OSTEOTOMIES} }

\nohyphens{Robert Grupp MS$^{1*}$, Yoshito Otake PhD$^{1,2}$, Ryan Murphy MSE$^{3,4}$, Javad Parvizi MD$^{5}$,\\Mehran Armand PhD$^{3,4}$, Russell Taylor PhD$^{1}$}

$^{1*}$Department of Computer Science, Johns Hopkins University, Baltimore MD, 21218, USA,\\grupp@jhu.edu

$^{2}$Graduate School of Information Science, Nara Institute of Science and Technology, Nara, 630-0192, Japan

$^{3}$Department of Mechanical Engineering, Johns Hopkins University, Baltimore MD, 21218, USA

$^{4}$Research and Exploratory Development Department, Johns Hopkins University Applied Physics Laboratory, Laurel, MD, 20723, USA

$^{5}$Rothman Institute, Thomas Jefferson University Hospital, Philadelphia PA 19107, USA\\

\textbf{INTRODUCTION}

The majority of pelvic osteotomy procedures are currently performed without acquiring preoperative CT scans.
For example, planning and diagnosis of Ganz periacetabular osteotomy (PAO) is usually performed using anteroposterior (AP) and false profile radiographs (Ganz 1988).
Surgeons must then rely on their own experience and intraoperative radiographs to perform the osteotomy and reposition the fragment (Ganz 1988).
The repositioning task is difficult for novice surgeons and the interpretation of intraoperative radiographs is difficult even for experienced practitioners (Troelsen 2009).
Computer-aided surgical (CAS) systems commonly use preoperative CT scans and optical systems to track tools when performing pelvic osteotomies (Langlotz 1997, 1998).
The Biomechanical Guidance System is an example CAS system developed to also track the osteotomized fragment and provide the surgeon with real-time updates of radigraphic angles for characterization of acetabulum and predicted contact pressure around the hip (Murphy 2014).
These CAS systems typically require full pelvic CT information; however, this is not desirable since most PAO patients are adolescents or females of childbearing age, and radiation exposure to the reproductive organs is a concern.
It has been shown that a partial CT of the patient, in conjunction with a statistical shape model (SSM) of the full pelvis, may be used to estimate the complete pelvis (Chintalapani 2010).
This complete estimation of the pelvis could allow the surgeon to gain the benefits of 3D planning and navigation while exposing the patient to a reduced dose of radiation required by the partial CT.
For these CAS systems to compute a registration transform between the patient's pelvic surface and the CT surface, points on the patient's ilium are digitized via the use of an optically tracked pointing tool (Murphy 2014).
Any errors in the estimation of the patient's complete pelvic surface via the SSM will propagate into the registration transform.

In this paper, we present the result of a smooth extrapolation technique that preserves known anatomy with no modification and extends Chintalapani's work  by providing a seamless transition to the estimated anatomy.
The constraints imposed by the smooth transition significantly reduce residual surface error.
We reduce radiation exposure by omitting CT slices both above and below the acetabulum, with the exception of a few slices of the superior iliac crest.
Using full female hip CTs, missing slices were simulated and a leave-one-out error analysis was conducted; we report on the accuracy of the two estimation methods.
%
\begin{figure}
        \centering
        \includegraphics[width=0.95\textwidth]{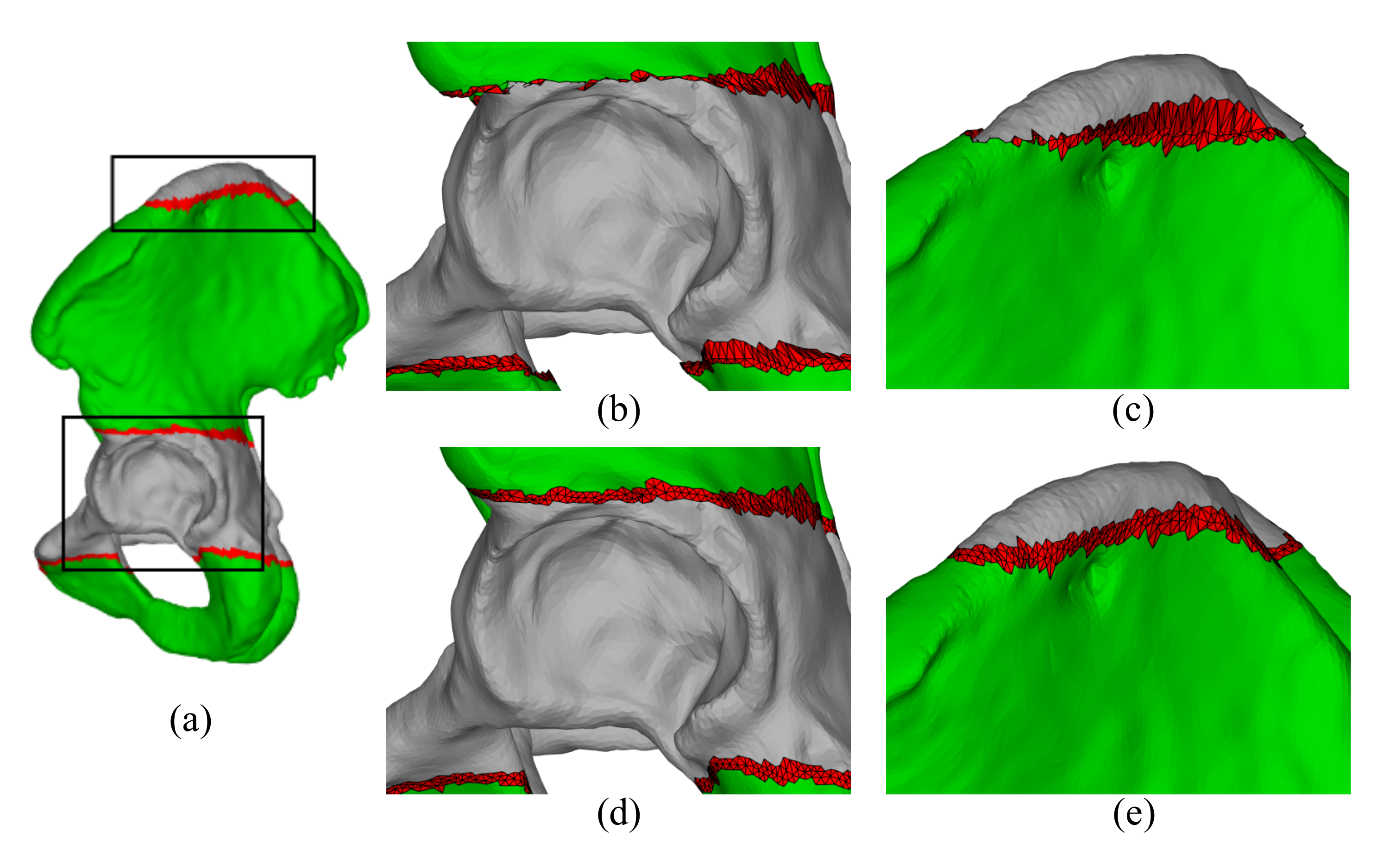}
        \caption{An example of extrapolation with the acetabulum and superior 5\% of the iliac crest used as priors.
        		     Grey regions indicate surfaces known from prior CT, green regions represent extrapolated regions, and red regions represent the boundary between the two.
		     (a) A lateral view of a pelvis, completed with smooth extrapolation.
		     (b) (c) Zoomed-in views of the same subject, but with the cut-and-paste extrapolation.
		     (d) (e) Zoomed-in views of the same subject with smooth extrapolation.
		     Note the continuous, smooth, transitions in the smooth case, in contrast to the discontinuities in the cut-and-paste case.
		     }
        \label{fig:extrap_example}
\end{figure}

%
\textbf{MATERIALS AND METHODS}

A SSM of the normal female pelvis was created using the pipeline described in (Grupp 2015).
Pelvis surfaces were modeled with triangular meshes generated by warping a single template mesh with displacement fields output from a deformable intensity-based volumetric registration.
The template mesh had been segmented beforehand and excluded the sacrum.
Initially, 70 hip CT volumes were considered, however the volumetric registration yielded 42 surface meshes of valid anatomy; this was determined by manual inspection.

A leave-one-out test was performed to evaluate the generalization capability of the SSM with complete pelvis anatomy used as prior.
During each iteration, one surface mesh is ``left out'' and a SSM is created from the remaining meshes.
The left out mesh is projected onto the modes of the SSM to obtain the optimal estimation of the left out mesh.
Surface error metrics are computed between the left out mesh and its estimate. 

Two extrapolation methods were used to estimate complete pelvis surfaces given an incomplete surface prior and a SSM.
Both methods use a partial surface prior as input to the SSM, which then produces an estimate of the complete anatomy.
The first method keeps the partial prior fixed and ``copies and pastes'' the remaining portion from the SSM estimate.
This process often results in a discontinuous transition from the known prior into the estimate, resulting in poor aesthetics and additional error (Chintalapani 2010).
The second, smooth method, keeps the partial surface prior fixed, but uses a Thin Plate Spline (TPS) and common regions between the partial prior and SSM estimate to model the transformation between the partial surface prior and SSM estimate (Grupp 2015).
The remaining unknown surface from the SSM estimate is transformed by the TPS and joined with the partial surface prior (Grupp 2015).
The exact interpolation between knot points of the TPS guarantees continuity between the boundary of the partial surface prior and unknown surface region from the SSM, while also ensuring smoothness  (Bookstein 1989).
\begin{figure}
        \centering
        \includegraphics[width=0.90\textwidth]{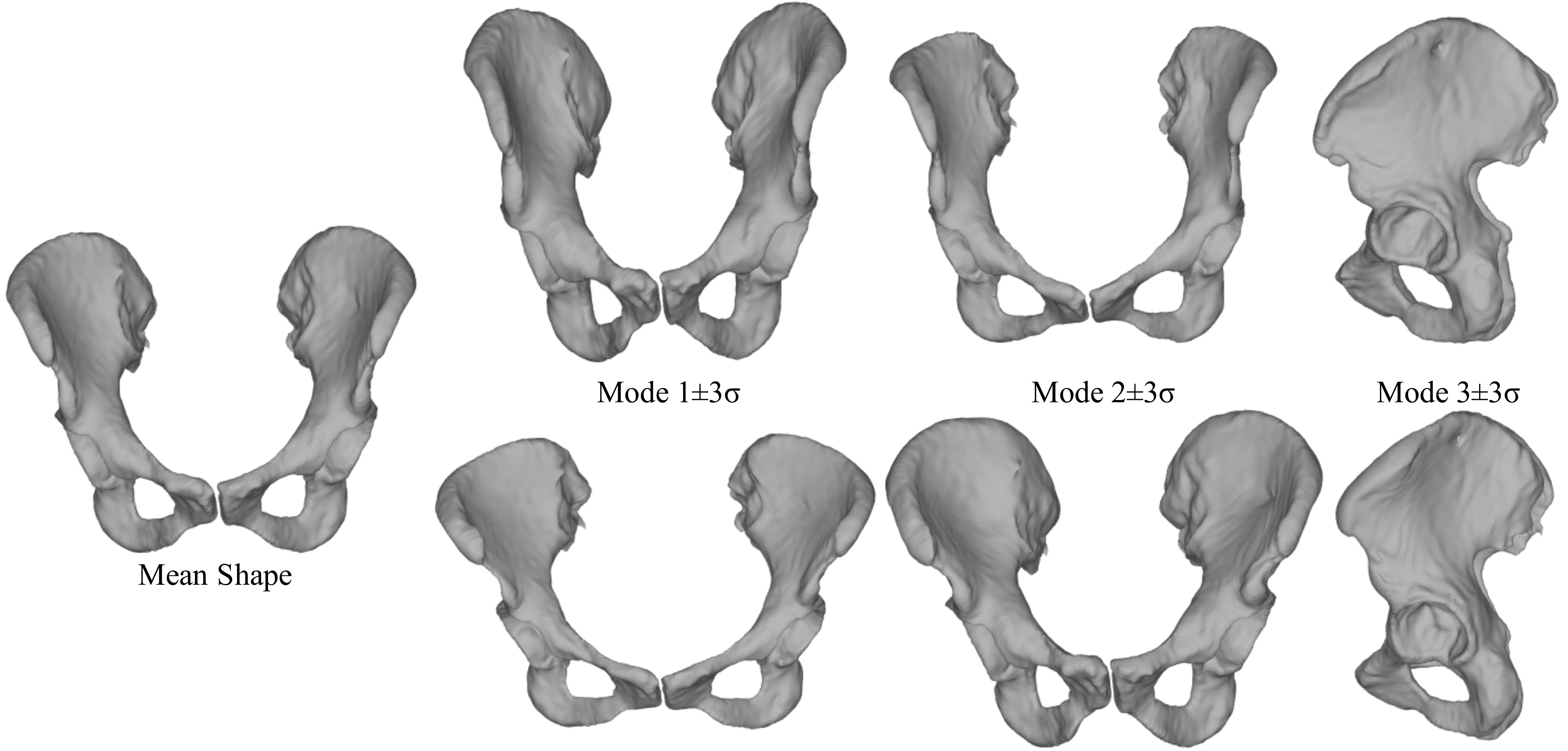}
        \caption{The mean shape and first three principle modes of the statistical model of the normal female pelvis.}
        \label{fig:normal_pelvis_modes}
\end{figure}

Another leave-one-out test was performed by simulating varying amounts of unknown anatomy, estimating the full anatomy, and computing surface error statistics in the extrapolated region.
All modes of the SSM were used during the extrapolation process.
The regions of known anatomy were defined as the axial slices that encompass the entire acetabulum and also superior axial slices of the iliac crest ranging from 0\% to 15\% of the entire pelvis height, in increments of 5\%. 
Figure \ref{fig:extrap_example} depicts a pelvis with the acetabulum and the superior 5\% of the iliac crest known and the remaining surface extrapolated using the SSM.

\textbf{RESULTS}

\begin{figure}
        \centering
        \begin{subfigure}[b]{0.49\textwidth}
                \includegraphics[width=\textwidth]{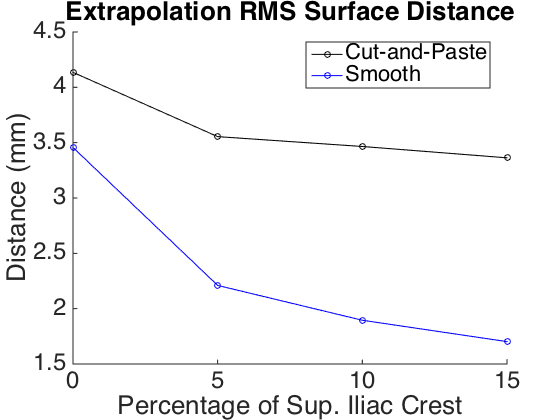}
                \caption{}
                \label{fig:rms_sur_dist_extrap}
        \end{subfigure}
        \begin{subfigure}[b]{0.49\textwidth}
                \includegraphics[width=\textwidth]{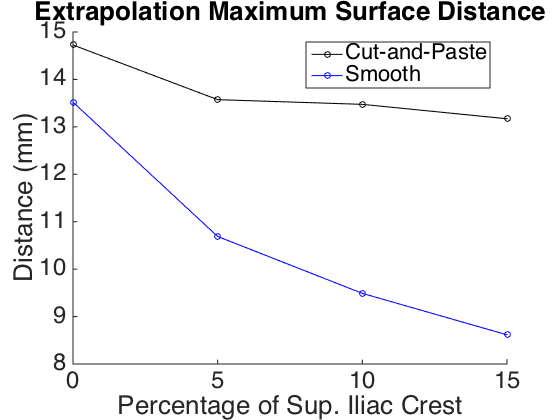}
                \caption{}
                \label{fig:max_sur_dist_extrap}
        \end{subfigure}
        \caption{Surface errors between true surfaces and extrapolated surfaces computed in the leave-one-out test.
        		    (a) RMS of mean surface errors.
		    (b) Average of maximum surface errors.} \label{fig:sur_dist}
\end{figure}

\begin{figure}
        \centering
        \includegraphics[width=0.95\textwidth]{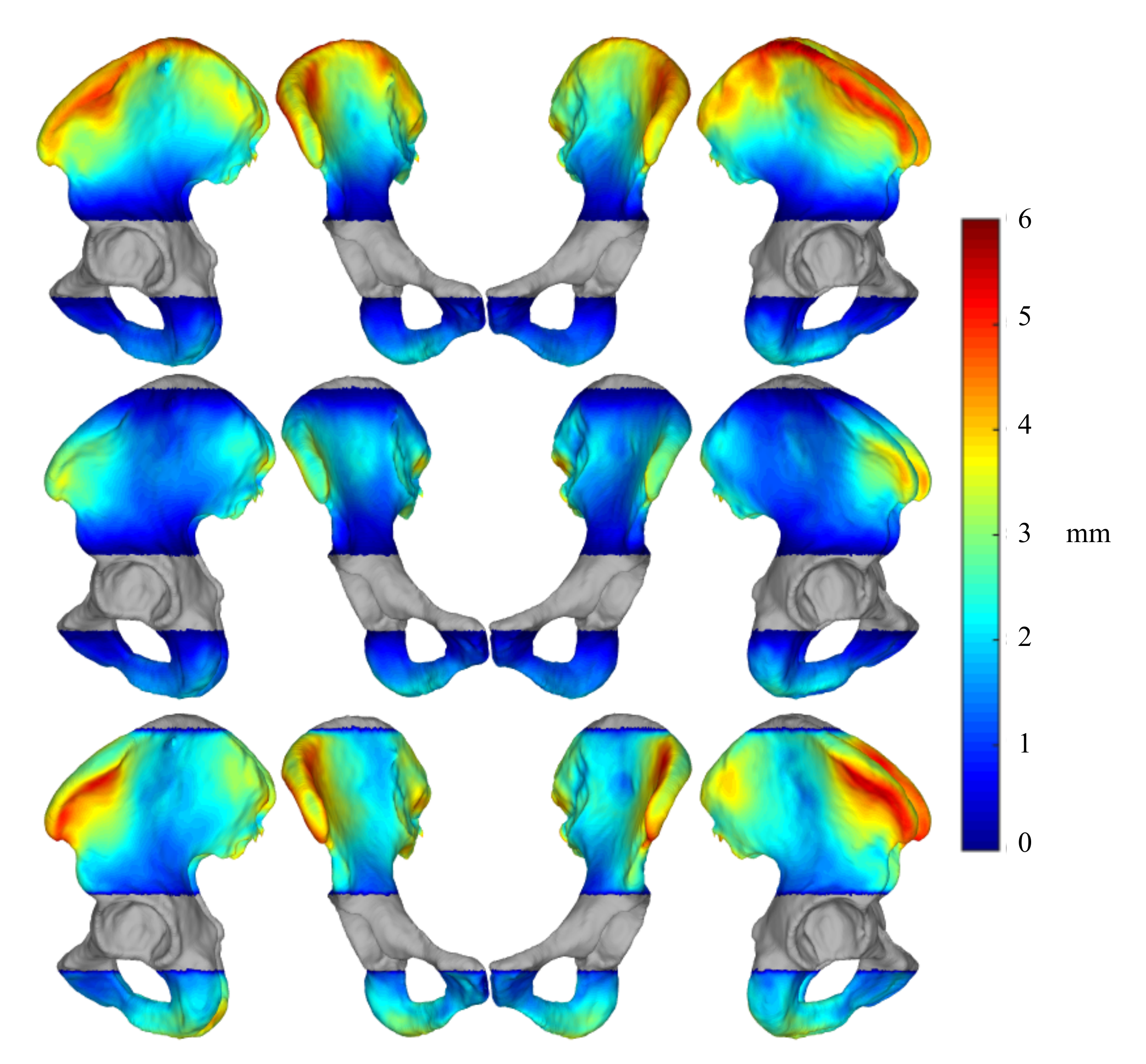}
        \caption{Mean surface errors present in extrapolated regions over the course of the leave-out experiment.
        		     The mean shape is used for visualization and grey regions indicate surfaces known from prior CT.
        		     Top: smooth extrapolation with only the acetabulum known as prior.
		     Center: smooth extrapolation with the acetabulum and top 5\% of the superior iliac crest known as prior.
		     Bottom: cut-and-paste extrapolation with the acetabulum and top 5\% of the superior iliac crest known as prior.
		     The additional constraint of the top 5\% of anatomy, in combination with the smooth extrapolation, reduces surface errors, whereas the cut-and-paste extrapolation is less effected.
		     }
        \label{fig:heat_map}
\end{figure}

The mean surface and first three principal modes ($\pm 3$ standard deviations) of a SSM constructed from all data are shown in Figure \ref{fig:normal_pelvis_modes}.
The SSM leave-one-out test conducted using complete anatomy of the pelvis showed a root mean square (RMS) surface error of 1.61 mm, a maximum surface error of 7.85 mm, and a RMS vertex error of 2.83 mm.
A specific example of the two extrapolation methods using a partial prior surface about the acetabulum and superior 5\% of the iliac crest is shown in Figure \ref{fig:extrap_example}.
The RMS and maximum surface errors computed using both extrapolation methods during the extrapolation leave-one-out test are shown in Figure \ref{fig:sur_dist}.
The smooth extrapolation method performed yielded an average improvement of 1.31 mm in RMS surface error over the cut-and-paste method and an average improvement of 3.16 mm in maximum surface error over the cut-and-paste method.
The minimum improvement in RMS and maximum surface error over the cut-and-paste approach was 0.68 mm and 1.21 mm, respectively, and coincided with the case of not retaining any of the superior iliac crest.
Figure \ref{fig:heat_map} highlights the distribution of surface errors over the course of the leave-one-out test for two different prior surface regions using the smooth approach, and for one prior surface region using the cut-and-paste approach.

%
\textbf{DISCUSSION}

We have shown that the smooth extrapolation method may be successfully applied to normal female pelvis anatomy and that it out-performs the cut-and-paste approach.
As shown by Figures \ref{fig:sur_dist} and \ref{fig:heat_map}, if a superior portion of the iliac crest is obtained as part of the preoperative partial CT, then surface error is significantly reduced when compared to the case with no superior portion of the iliac crest.
When using complete anatomy as prior, the SSM has 1.61 mm RMS surface error, compared with 1.70 mm, 1.90 mm, 2.21 mm, and 3.50 mm RMS surface error (Figure \ref{fig:sur_dist}) in the extrapolated regions when estimating using the smooth extrapolation method and a prior consisting of the acetabulum and 15\%, 10\%, 5\%, and 0\% of the superior iliac crest.
Therefore, with a sufficient non-zero amount of the superior iliac crest used as prior, the smooth extrapolation method creates less than 1 mm of additional surface error to the generalization error of the SSM.

If used as input to a CAS system, smoothly extrapolated pelvis surfaces could result in more accurate patient-to-CT registration transformations than those computed using extrapolated surfaces via the cut-and-paste approach, especially when considering 2D/3D image-based registration (Otake 2012).
By reducing radiation exposure for the patient and maintaining navigational accuracy, computer-aided pelvic osteotomies have the capability to improve surgical outcomes for those suffering from dysplasia.

Future work will incorporate intraoperative radiographs to the extrapolation process, simulation of a CAS system's typical patient-to-CT registration using smoothly extrapolated surfaces as input, and creating a more robust SSM.
Moreover, construction of a SSM from dysplastic pelves has direct application to navigating PAO with a system such as the BGS (Murphy 2014).
Intraoperative radiographs would help to further reduce the uncertainty associated with the SSM estimates derived from partial surfaces.
With the creation of a robust dysplastic pelvis SSM, we hope to achieve an estimation method that has negligible errors and be realistically applied to relevant computer-assisted pelvic osteotomies.

\textbf{REFERENCES}

\vspace*{-12pt} 
\begin{itemize}
	\item Bookstein F, Principal warps: thin-plate splines and the decomposition of deformations, IEEE Transactions on Pattern Analysis and Machine Intelligence, 11(6), pp: 567-585, 1989.
	\item Chintalapani G, Statistical atlases of bone anatomy and their applications, Johns Hopkins University, 2010.
	\item Chintalapani G, Murphy R, Armiger R, Lepist{\"o} J, Otake Y, Sugano N, Taylor R, Armand M,  Statistical atlas based extrapolation of CT data, SPIE Medical Imaging, pp: 762539-762539, 2010.
	\item Ganz R, Klaue K, Vinh T, Mast J, A new periacetabular osteotomy for the treatment of hip dysplasias Technique and preliminary results, Clinical orthopaedics and related research 232, pp: 26-36, 1988.
	\item Grupp R, Chiang H, Otake Y, Murhpy R, Gordon C, Armand M, Taylor R, Smooth extrapolation of unknown anatomy via statistical shape models, SPIE Medical Imaging, 2015 (to appear).
	\item Langlotz F, Stucki M, B{\"a}chler R, Scheer C, Ganz R, Berlemann U, Nolte L, The first twelve cases of computer assisted periacetabular osteotomy, Computer Aided Surgery, 2(6), pp: 317-326, 1997.
	\item Langlotz F, B{\"a}chler R, Berlemann U, Nolte L, Ganz R, Computer assistance for pelvic osteotomies, Clinical orthopaedics and related research, 354, pp: 92-102, 1998.
	\item Murphy R, Armiger R, Lepist{\"o} J, Mears S, Taylor R, Armand M, Development of a biomechanical guidance system for periacetabular osteotomy, International journal of computer assisted radiology and surgery, pp: 1-12, 2014.
	\item Otake Y, Armand M, Armiger R, Kutzer M, Basafa E, Kazanzides P, Taylor R, Intraoperative image-based multiview 2D/3D registration for image-guided orthopaedic surgery: incorporation of fiducial-based C-arm tracking and GPU-acceleration, IEEE Transactions on Medical Imaging, 31(4), pp: 948-962, 2012.
	\item Troelsen A, Surgical advances in periacetabular osteotomy for treatment of hip dysplasia in adults, Acta Orthopaedica, 80(s332), pp: 1-33, 2009.
\end{itemize}

\textbf{DISCLOSURES}

None to report.

\end{document}